\documentclass{article}
\usepackage{spconf,amsmath,graphicx, multirow}
\usepackage{booktabs}

\title{Speaker Adaptation for End-to-End CTC Models}
\name{Ke Li$^{1,2}$ \thanks{Ke Li performed the work while she was a research intern at Microsoft
AI and Research}, Jinyu Li$^{1}$, Yong Zhao$^{1}$, Kshitiz Kumar$^{1}$, Yifan Gong$^{1}$}
\address{$^{1}$ Microsoft AI and Research
		\\$^{2}$ Center for Language and Speech Processing, Johns Hopkins University}
\begin{document}
\ninept
\maketitle
\begin{abstract}
We propose two approaches for speaker adaptation in end-to-end (E2E) automatic speech recognition systems. One is Kullback-Leibler divergence (KLD) regularization and the other is multi-task learning (MTL). Both approaches aim to address the data sparsity especially output target sparsity issue of speaker adaptation in E2E systems. The KLD regularization adapts a model by forcing the output distribution from the adapted model to be close to the unadapted one. The MTL utilizes a jointly trained auxiliary task to improve the performance of the main task. We investigated our approaches on E2E connectionist temporal classification (CTC) models with three different types of output units. Experiments on the Microsoft short message dictation task demonstrated that MTL outperforms KLD regularization. In particular, the MTL adaptation obtained 8.8\% and 4.0\% relative word error rate reductions (WERRs) for supervised and unsupervised adaptations for the word CTC model, and 9.6\% and 3.8\% relative WERRs for the mix-unit CTC model, respectively.

\end{abstract}
\section{Introduction}
\label{sec:intro}
End-to-end (E2E) systems are an emerging field in automatic speech recognition (ASR) \cite{
miao2015eesen,Chan-LAS,soltau2016neural,prabhavalkar2017comparison,battenberg2017exploring,sak2017recurrent,chiu2018state,sainath2018improving,Li18CTCnoOOV,audhkhasi2017direct, chen2018modular}. An E2E system directly transduces an input sequence of acoustic features to an output sequence of tokens; it works effectively since ASR is inherently a sequence-to-sequence task, mapping an input waveform to an output token sequence. Prominent E2E approaches include: (a) connectionist temporal classification (CTC) \cite{Graves-CTCFirst,Graves-E2EASR}, (b) attention based encoder-decoder networks \cite{Cho-RNNEncDecSMT,Bahdanau-RNNEncDecAlignTranslate,Bahdanau-AttentionASR,Chorowski-AttentionASR}, and (c) recurrent neural network (RNN) transducer \cite{Graves-RNNSeqTransduction}. The above approaches have been successfully applied to large scale ASR tasks \cite{miao2015eesen,Chan-LAS,soltau2016neural,prabhavalkar2017comparison,battenberg2017exploring,chiu2018state,Li18CTCnoOOV,sak2015fast,rao2017exploring,audhkhasi2017direct,Li17CTCnoOOV,Das18CTCAttention,chen2016phone,chen2017confidence}. 

The E2E systems are typically trained on large scale data from millions of users, and have shown comparable performance with the state-of-the-art deep network based hybrid systems \cite{soltau2016neural, chiu2018state,Li18CTCnoOOV}. Given the fast development of E2E systems, it is worthwhile to investigate how to adapt E2E systems to new speakers with limited adaptation utterances. In this study, we use E2E CTC systems as the platform to study general speaker adaptation technologies, which should be able to generalize to other types of E2E systems.

Various approaches have been proposed for speaker adaptation in hybrid systems. These methods can be classified into three categories. First approach is inherently fine-tuning. Certain layers or the whole speaker-independent (SI) model are updated \cite{liao2013speaker,yu2013kl}. To avoid overfitting, L2 regularization using weight decay \cite{liao2013speaker} and Kullback-Leibler divergence (KLD) regularization \cite{yu2013kl} were proposed. The second category of approaches is transformation based. The basic assumption is that there exists a linear transformation that can convert a SI model to a speaker-dependent (SD) model. This approach inserts a SD linear transformation layer into the SI model~\cite{seide-2011,Xue14,swietojanski2016learning,zhao2015,huang2015rapid,delcroix2015context,wu2015multi,Lahiru17}. Third, speaker aware training with i-vectors~\cite{Saon13,miao2014towards}, speaker code~\cite{Abdel13}, and speaker embedding~\cite{cui2017embedding}  were proposed to facilitate the system to normalize the speaker variations. Besides, a linear least square method is utilized for unsupervised adaptation for deep neural networks (DNNs)~\cite{hsiao2015unsupervised}. 

The aforementioned adaptation methods are mainly investigated in the traditional hybrid systems. Though these methods can be applied regardless of the underlying SI models, it's effectiveness for adapting E2E systems is uncertain. There are three challenges of the speaker adaptation in E2E CTC systems. 
First, E2E CTC models with word targets have been shown to significantly outperform those with subword or letter units \cite{soltau2016neural, Li18CTCnoOOV}. However the speaker adaptation of such word CTC systems confronts with severe output target sparsity issue because the number of the output units is typically tens of thousands. Thus most word units do not appear in the limited adaptation data, and it's hard to guarantee the adaptation towards the speakers, instead of overfitting the limited word targets.
Second, when the acoustic modeling technology evolves from DNNs to RNNs with long short-term memory (LSTM) units widely used in both hybrid and E2E systems, the difficulty in improving the adaptation performance of LSTMs has increased~\cite{miao2015,zhao2017}. For example, ~\cite{miao2015} reported only around 4\% relative word error rate (WER) reduction, partially because the recurrent topology of the LSTMs makes it more effective to capture and normalize long-range speaker characteristics than the DNNs. 
Moreover, compared to the traditional DNN and LSTM acoustic models in hybrid systems, the adaptation of CTC models is expected to mitigate the mismatch in both acoustic and linguistic conditions, making it more challenging than adapting hybrid systems. 

There have been few efforts in the literature on the adaptation of the E2E systems. Recently, ~\cite{mirsamadi2017multi} investigated domain adaptation for CTC models by inserting a linear transformation layer after the first hidden layer and learning it on 10 hours transcribed adaptation data with the standard CTC criterion for adaptation. Given enough adaptation data, data sparsity is not an issue for such domain adaptation.  ~\cite{ochiai2018speaker} proposed a multi-path adaptation scheme to improve ASR performance in the attention based encoder-decoder system and observed that adapting the encoder network is more effective than adapting other components. 

In this work, we address the data sparsity issue by formulating KLD regularization and multi-task learning (MTL) approaches for speaker adaptation of E2E CTC models. We apply them on three E2E CTC models with different output targets: letters, words, and mix-units ~\cite{Li17CTCnoOOV,Li18CTCnoOOV}. The KLD regularization constrains the SD model to be close to the SI model, and potentially avoids overfitting to limited speaker adaptation data. This is especially critical for unsupervised adaptation where we use decoded hypotheses as references. However, the KLD regularization cannot address the target sparsity issue in the limited adaptation data when the networks have tens of thousands of output units. As a solution, we propose MTL adaptation that uses an auxiliary task with fewer output units to improve the performance of major task with large output units in E2E CTC models. We jointly optimize the CTC losses of both tasks by adjusting the shared hidden layers on adaptation data. Almost all units in the auxiliary task can be observed in very limited adaptation data. Thus the auxiliary task addresses the output sparsity issue and likely guides adaptation towards target speakers. 
The rest of the paper is organized as follows. In Sec.~\ref{sec:Adapt}, we propose and discuss KLD and MTL-based speaker adaptations for E2E CTC models. We evaluate the proposed adaptation techniques in Sec.~\ref{sec:exp}, and present conclusions in Sec.~\ref{sec:con}.

\section{Speaker Adaptation for E2E CTC Systems}
\label{sec:Adapt}

\subsection{E2E CTC Models}
\label{ssec: CTC}
We use LSTM-RNNs as model architectures and CTC loss as objective function~\cite{Graves-CTCFirst, Graves-E2EASR} to optimize the prediction of a transcription sequence. The ASR output symbols in an utterance are usually fewer than the input speech frames. Hence CTC paths are used to force the output to have the same length as the input speech frames by adding blank as an additional label and allowing repetition of labels. Let us denote $\bf{x}$ as the speech input sequence, $\Theta$ as the network parameters, $\pi$ as the CTC path, $\bf{l}$ as the original label sequence, and  $B^{-1}(\bf{l})$ as all possible CTC paths expanded from $\bf{l}$. 

The CTC loss function is defined as the sum of negative log probabilities of correct labels as:
\begin{equation}
L_{CTC} = - \ln P_{\Theta}( {\bf{l}|\bf{x}} ) = - \ln \sum_{{\pi} \in B^{-1}(\bf{l})} P_{\Theta}( {\pi}  | \bf{x} )
\label{eq:ctc}
\end{equation}
Based on the conditional independence assumption for output units, $P( {\pi} | \bf{x} )$ can be decomposed to a product of posteriors from each time step $t$ as below:
\begin{equation}
P_{\Theta}( {{\pi}  | \bf{x}} ) = \prod_{t=1}^T P_{\Theta}( \pi_{t}| x_t),
\end{equation}
where $x_t$ is the input speech frame at time $t$, $\pi_{t}$ is the output unit at time $t$, and $T$ is the sequence length in frames.

CTC outputs are usually dominated by blank labels. The outputs corresponding to the non-blank labels usually occur with spikes in their posteriors. Thus, an easy way to generate ASR outputs using CTC is to concatenate the non-blank labels corresponding to the posterior spikes and collapse those labels into word outputs if needed~\cite{soltau2016neural,Graves-CTCFirst}. This is known as greedy decoding; it is a very attractive feature for E2E modeling as it doesn't require language model (LM) or complex decoding. We too use greedy decoding in this study.

The output labels of E2E CTC systems can be either letters or words. As the goal of ASR is to generate a word sequence from the speech waveform, word units are the most natural output units. As shown in~\cite{soltau2016neural, Li18CTCnoOOV}, E2E CTC models with word targets significantly outperform those with letter units. A big challenge in the word CTC model, however, is the out-of-vocabulary (OOV) issue since it is infeasible to model all words in the network's output layer. Only the most frequent words in the training set are used as output targets whereas remaining words are tagged as OOVs. In~\cite{Li18CTCnoOOV}, a solution was proposed by decomposing a word sequence with OOVs into a mix-unit sequence of frequent words and characters. Using mix-units as output targets not only solves the OOV issue but also improves the recognition accuracy.  

In this paper, we will investigate the speaker adaptation technologies for E2E CTC models with three types of output units: letters, words, and mix-units. The output target sequences for supervised and unsupervised adaptations are ground-truth transcriptions and decoding hypotheses, respectively. The simplest way of speaker adaptation is to fine tuning the SI model with speaker adaptation data using Eq. \eqref{eq:ctc}.  
Compared to letter units, the number of words or mix-units is orders of magnitude larger. This makes the data sparsity issue increasingly challenging.

\subsection{Kullback-Leibler divergence (KLD) Adaptation}
To avoid overfitting, we first propose to regularize the CTC objective with a KLD metric between the posterior distributions of the adapted and unadapted models, which constrains the SD model ${\Theta}^s$ to not deviate far from that of the SI model $\Theta$. The KLD metric is:
\begin{equation}
	D_{KL} = \sum_{t=1}^T \sum_{u \in U}  P_{\Theta}( \pi_{t}=u| x_t) \ln \left( \frac{P_{\Theta}( \pi_{t}=u| x_t)}{P_{{\Theta}^s}( \pi_{t}=u| x_t)} \right),
	\label{eq:KL}
\end{equation}
where $u$ is an output unit of the CTC model, and $U$ is the corresponding token list of output units.
We add the KLD regularization into Eq.~\eqref{eq:ctc} and obtain a regularized objective function: 
\begin{align}
L_{KL\_CTC} = &(1-\alpha)  L_{CTC} + \alpha D_{KL} 	\nonumber \\
						 = &- (1-\alpha) \ln \sum_{{\pi} \in B^{-1}(\bf{l})}  \prod_{t=1}^T P_{{\Theta}^s}( \pi_{t}| x_t) \nonumber \\
								& + \alpha \sum_{t=1}^T \sum_{u \in U}  P_{\Theta}( \pi_{t}=u| x_t) \ln \left( \frac{P_{\Theta}( \pi_{t}=u| x_t)}{P_{{\Theta}^s}( \pi_{t}=u| x_t)} \right),
\label{eq:KLCTC1}
\end{align}
where $\alpha$ is the KLD regularization coefficient and its range is [0,1]. $\alpha = 0$ reduces the objective to the default CTC criterion without regularization. 
Because $P_{\Theta}( \pi_{t}=u| x_t) \ln P_{\Theta}( \pi_{t}=u| x_t) $ does not affect the optimization of the SD model, we rewrite the objective function as:
\begin{align}
	 L_{KL\_CTC} = & - (1-\alpha) \ln \sum_{{\pi} \in B^{-1}(\bf{l})}  \prod_{t=1}^T P_{{\Theta}^s}( \pi_{t}| x_t) \nonumber \\
								& - \alpha \sum_{t=1}^T \sum_{u \in U}  P_{\Theta}( \pi_{t}=u| x_t) \ln P_{{\Theta}^s}( \pi_{t}=u| x_t).
	\label{eq:KLCTC}
\end{align}

In this study, we investigate KLD adaptation for CTC models by updating: 1) all layers (All), 2) all hidden layers with the top softmax layer fixed (Hidden), and 3) only the top softmax layer (Top). 

\subsection{Multi-task Learning (MTL) Adaptation}
\label{ssec:MTL}
The KLD CTC adaptation prevents the adapted SD model from deviating far from the SI model. However, the regularization does not eliminate the output target sparsity issue when the E2E CTC models use a large number of words or mix-units as output targets, where only a very small portion of which can be observed during adaptation. The learning procedure tends to improve the criterion by simply increasing the probabilities of observed targets and pushing the probabilities of unseen targets towards zero, which is not optimal. This issue is exacerbated in unsupervised adaptation, where the target output sequence may inevitably contain errors. 

Inspired by the MTL adaption work in \cite{huang2015rapid} that adapts senone-based hybrid models with an auxiliary monophone target, we propose MTL adaptation of CTC models to better address the data sparsity issue. In addition to the word (or mix-unit) classification as the primary task, an auxiliary task is introduced to conduct letter classification. This yields an MTL network that predicts both words (or mix-units) and letters with shared hidden layers. The number of letters is typically around 30, and the letter units are usually efficiently covered in the limited adaptation data. The CTC criterion for the letter classification task is:
\begin{equation}
L_{CTC-l}  = - \ln \sum_{{\phi} \in B^{-1}(\bf{m})}  \prod_{t=1}^T P_{{\Theta}}( \phi_{t}| x_t),
\label{eq:ctc-letter}
\end{equation}
where $\bf{m}$ is the letter sequence, and  $B^{-1}(\bf{m})$ denotes all possible CTC paths expanded from $\bf{m}$. 

Finally, we create the MTL CTC adaptation criterion as a linear combination of the CTC losses from word (or mix-unit) and letter targets:
\begin{align}
L_{MTL\_CTC} = &(1-\beta)  L_{CTC} + \beta L_{CTC-l} 	\nonumber \\
						 = &- (1-\beta) \ln \sum_{{\pi} \in B^{-1}(\bf{l})}  \prod_{t=1}^T P_{{\Theta}^s}( \pi_{t}| x_t) \nonumber \\
								& - \beta  \ln \sum_{{\phi} \in B^{-1}(\bf{m})}  \prod_{t=1}^T P_{{\Theta}^s}( \phi_{t}| x_t),
\label{eq:MTLCTC}
\end{align}
where $\beta$ is the weight for the letter CTC model. $\bf{l}$ in Eq. \eqref{eq:MTLCTC} indicates either word or mix-unit sequence. 

The procedure of the MTL adaptation for word (or mix-unit) CTC models is as follows:
\begin{itemize}
	\item Insert a softmax layer with the letter targets on top of the SI model's last hidden layer. 
	\item Train the letter softmax layer with all training data to predict the letters using Eq.~\eqref{eq:ctc-letter}, keeping the parameters of the original E2E CTC model fixed.
	\item For each speaker, update the shared hidden layers on adaptation data using Eq.~\eqref{eq:MTLCTC}, keeping both softmax layers fixed.
    \item During decoding of the adapted speaker model, we just evaluate word (or mix-unit) outputs and ignore evaluating letter outputs.
\end{itemize}

The interpolation weight $\beta$ ranges from 0 to 1. When $\beta = 0$, we update the hidden layers with the default word (or mix-unit) based CTC criterion. When $\beta = 1$, the hidden layers are updated with purely the letter CTC criterion although the model later will be used to generate word (or mix-unit) output sequences. The difference between word and mix-unit models for unsupervised MTL adaptation are the letter targets. For the word model, the letter targets are from decoding the adaptation data with the letter branch. The decoded results from the word branch can not be decomposed into letter targets since it contains OOVs. While for the mix-unit model, considering it generates more accurate hypotheses than the letter branch and does not have the OOV issue, the letter targets in MTL are generated by decomposing the decoding hypotheses of the adaptation data from the mix-unit branch. 



\section{Experiments and Results}
\label{sec:exp}

We evaluate our proposed speaker adaptation methods for the E2E CTC systems on a Microsoft short message diction (SMD) task. 

\subsection{Speaker Independent Model Setup}
\label{ssec:SIsetup}

Using around 3400 hours Microsoft US-English Cortana utterances, we trained three E2E CTC SI models with letters, words, and mix-units as output targets, respectively. Above SI models are a 6-layer bi-directional LSTM, with 512 memory units per layer in each direction. We derived 80-dimensional log Mel filterbank energies at 10-ms intervals and stacked 3 contiguous frames to form 240-dimensional features for CTC \cite{sak2015fast}. For the letter CTC model, the size of output is 30 including 26 English characters [a-z], ', *, \$, and a blank symbol. \$ denotes space which means the boundary of two words.  For the word CTC model, the output contains about 27k frequent words (occurred at least 10 times in the training data) and the rest words are mapped to an OOV token. The vocabulary size of the  mix-unit CTC model is about 33k, consisting of both frequent words and letter ngrams \cite{Li18CTCnoOOV}. Units between two \$ form one word. We use greedy decoding  by simply selecting non-blank units with the highest posteriors and collapse them into word sequences.  

We have two test sets. One contains 7 speakers  with 200 utterances per speaker and a total of 20k words. The other is a 50-speaker set with 182k words. We first study speaker adaptation for the E2E CTC models on the 7-speaker dataset for a quick assessment, and then pick the best setup to evaluate the adaptation performance on the 50-speaker test set. The E2E CTC SI models with letters, words, and mix-units targets have respectively 28.44\%, 17.24\%, and  16.76\% WERs on the 7-speaker test set, and respectively 23.14\%, 13.88\%, and 13.63\% WERs on the 50-speaker test set. The performance gap between the 7 and 50-speaker set is due to a speaker with significantly higher WER in the 7-speaker set.  We evaluate the performance under both supervised and unsupervised conditions with 10, 50, and 200 utterances per speaker. The adaptation and test sets are separated. 

\subsection{Speaker Adaptation for CTC with Letter Outputs}
For the letter CTC model, we simply apply the KLD adaptation as it has only 30 output targets and thus does not have the adaptation target sparsity issue. The results of supervised KLD adaptation with three parameter updating setups on the 7-speaker test set are in Table~\ref{tab:letter-sup}. All three updating setups obtain minor improvements using the smallest adaptation set with 10 utterances per speaker. For adapting with 50 and 200  utterances scenarios, both adapting all layers and all hidden layers improve the baseline letter CTC model and perform better than adapting only the top softmax layer. Among the three setups, adapting only hidden layers gives the best WER 24.89\% and relative word error rate reduction (WERR) 12.5\% with 200 adaptation utterances and $\alpha=0.0$. 
We consistently observe the least improvements from adapting only top layer in the following experiments, thus we do not report any results with such setup to save space.

We present results of unsupervised KLD adaptation with the same 7-speaker test dataset in Table~\ref{tab:letter-uns}. The observations for unsupervised scenario are consistent with the supervised case except for smaller relative WERRs. The best WERR is 2.25\% by adapting all hidden layers on 200 adaptation utterances with $\alpha=0.2$. 


\begin{table}[t]
  \caption{WERs (\%)  on the 7-speaker set for supervised KLD regularization adaptation for the letter CTC model. 
  }
  \label{tab:letter-sup}
  \centering
  \begin{tabular}{l|c|c|c|c|c}
    \toprule
		Adapt. type 		& $\#$ of utt. & 	$\alpha$=0		& 	$\alpha$=0.2 	&		$\alpha$=0.5 	& 	$\alpha$=0.8	\\
   	\midrule
        Baseline &  - & \multicolumn{4}{c}{28.44} \\
    \midrule
										&		10				& 	28.79 & 	28.36		& 		28.18	& 28.17 \\
		All							&		50				& 	27.47 & 27.53	& 	27.60	& 28.03 \\
										&		200				& 	\textbf{25.15} & 26.09 & 	26.85	& 27.65 \\	\hline
										&		10				& 	28.32 & 	28.26		& 		28.19	& 28.20 \\
		Hidden					&		50				& 	27.06 & 	27.50	& 	27.66	& 28.09 \\
										&		200				& 	\textbf{24.89} & 25.99	& 	26.87	& 27.64 \\	\hline
										&		10				& 	29.40 & 	28.63	& 	28.26	& 28.41 \\
		Top							&		50				& 	29.58 & 	28.51	& 	28.25	& 28.36 \\
										&		200				& 	28.69 & 	28.11	& 	\textbf{28.08}	& 28.25 \\												
    \bottomrule
  \end{tabular}
\end{table}

\begin{table}[t]
  \caption{WERs (\%) on the 7-speaker set for unsupervised KLD regularization adaptation 
  for the letter CTC model.}
  \label{tab:letter-uns}
  \centering
  \begin{tabular}{l|c|c|c|c|c}
    \toprule
		Adapt. type 		& $\#$ of utt. & 	$\alpha$=0		& 	$\alpha$=0.2 	&		$\alpha$=0.5 	& 	$\alpha$=0.8	\\
   	\midrule
        Baseline &  - & \multicolumn{4}{c}{28.44} \\
    \midrule
										&		10				& 	28.50 & 	28.29		& 		28.38	& 28.32 \\
		All							&		50				& 	28.30 & 	28.10		& 		28.20	& 28.22 \\
										&		200				& 	28.05 & \textbf{27.91}& 		28.06	& 27.98 \\	\hline
										&		10				& 	28.18 & 	28.22		& 		28.38	& 28.22 \\
		Hidden					&		50				& 	28.01 & 	28.11		& 		28.07	& 28.24 \\
										&		200				& 	27.83 & 	\textbf{27.80}		& 		28.01	&  27.99\\	
    \bottomrule
  \end{tabular}
\end{table}

\begin{table}[t]
  \caption{WERs and relative WERR of supervised and unsupervised KLD adaptation on the 50-speakers set for the letter CTC model.}
  \label{tab:letter-50spk}
  \centering
  \begin{tabular}{l|c|c|c}
    \toprule
								 		& $\#$ of utt. & 	WER (\%)		& relative WERR	(\%) \\
    \midrule
		Baseline						&		-				& 	23.14 & 	-	  \\	\hline
										&		10				& 	22.91 & 	1.0	 \\
		Supervised						&		50				& 	21.85 & 	5.6	\\
										&		200				& 	\textbf{20.30} & \textbf{12.3}	 \\	\hline
										&		10				& 	23.14 & 	0.0		\\
		Unsupervised					&		50				& 	23.00 & 	0.6		\\
										&		200				& 	\textbf{22.77} & \textbf{1.6}		\\	
    \bottomrule
  \end{tabular}
\end{table}

Based on the best settings of the KLD adaptation on the 7-speaker dataset, namely tuning hidden layers with $\alpha=0.0$ for supervised case and $\alpha=0.2$ for the unsupervised case, we evaluate the performances on a 50 speaker dataset.  
We can see from Table~\ref{tab:letter-50spk} that the supervised adaptation on 200 utterances provides up to 12.3\% relative WERR while the unsupervised adaptation on the same data obtains at most 1.6\%. The results on the 50-speaker dataset are consistent with those on the 7-speaker dataset and indicate that unsupervised adaptation is more challenging than the supervised case.

\subsection{Speaker Adaptation for CTC with Word Outputs}
For the word CTC model, we apply both the KLD and MTL adaptations. The WERs of the SI word CTC model are 17.24\%, and 13.88\% on the 7-speaker and the 50-speaker datasets, respectively. 
\subsubsection{KLD Adaptation}
\begin{figure}[t]
\centering
\resizebox{1.0\linewidth}{!}{
\includegraphics{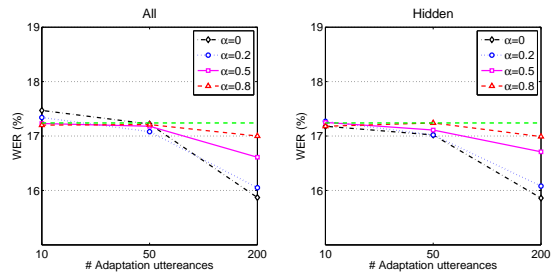}
}
\caption{WERs(\%) for the supervised KLD adaptation of the word CTC model on the 7-speaker dataset. The left and right sub-figures respectively present the results for adapting all network layers and all hidden layers.}
\label{fig:KL}
\end{figure}

We show results of supervised KLD adaptation on the 7-speaker dataset in Fig.~\ref{fig:KL}. The left sub-figure shows results by adapting all layers. Adapting with 10 utterances per speaker does not show obvious improvements regardless of values of $\alpha$. There is a WER regression without regularization, \emph{i.~e.} $\alpha=0$, indicating that adaptation with only CTC criterion is likely overfitted. For 50 utterances per speaker scenario, we see a strong relevance for regularization and obtain the best WER with $\alpha=0.2$. Furthermore, with 200 utterances we achieve the best WER (15.87\%) with $\alpha=0$, consistent with the result of the letter model. 
We report very similar observations when adapting all hidden layers in the right sub-figure in Fig.~\ref{fig:KL}. 

We report results of unsupervised KLD adaptation in Table~\ref{tab:wer}. We obtain decoded hypotheses from the SI word CTC model on adaptation data, and use them as targets in adaptation. The hypotheses inevitably have errors, requiring better regularization to prevent overfitting towards incorrect targets. Adaptation without regularization on 10 utterances regresses. The unsupervised KLD adaptation shows modest improvements compared with supervised scenario. The best WER is 16.92\%, or 1.9\% relative WERR over the baseline SI word CTC model, by adapting all the layers on 200 utterances with $\alpha=0.2$.

\begin{table}
  \caption{WERs (\%) of unsupervised KLD regularization adaptation for the word CTC model on the 7-speaker set.
  .}
  \label{tab:wer}
  \centering
  \begin{tabular}{l|c|c|c|c|c}
    \toprule
		Adapt. type 		& $\#$ of utt. & 	$\alpha$=0		& 	$\alpha$=0.2 	&		$\alpha$=0.5 	& 	$\alpha$=0.8	\\
    \midrule
        Baseline &  - & \multicolumn{4}{c}{17.24} \\
    \midrule
										&		10				& 	17.53 & 	17.38		& 		17.30	& 17.22 \\
		All							&		50				& 	17.56 & 	17.38		& 		17.31	& 17.26 \\
										&		200				& 	17.05 & \textbf{16.92}& 		17.09	& 17.17 \\	\hline
										&		10				& 	17.32 & 	17.24		& 		17.21	& 17.24 \\
		Hidden					&		50				& 	17.29 & 	17.30		& 		17.32	& 17.28 \\
										&		200				& 	17.04 & 	\textbf{17.01}		& 		17.18	& 17.19 \\	
    \bottomrule
  \end{tabular}
\end{table}

\begin{figure}
\centering
\resizebox{1.0\linewidth}{!}
{
\includegraphics{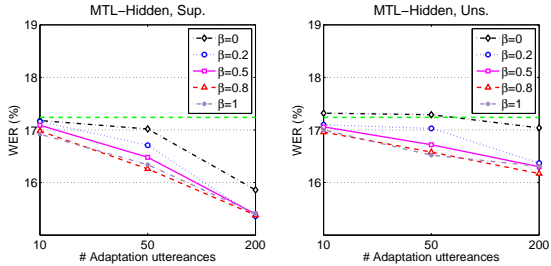}
}
\caption{WERs (\%) for the MTL adaptation of the word CTC model on the 7-speaker dataset. The left and right sub-figures present supervised and unsupervised adaptations, respectively.}
\label{fig:MTL}
\end{figure}

\subsubsection{MTL Adaptation}
Although KLD adaptation obtains modest improvements, it does not fully resolve the output target sparsity issue. For example, the average number of words per utterance is around 15. Even with 200 adaptation utterances, we can only observe 3k word targets, much smaller than the 27k vocabulary in the word CTC model. This target sparsity issue can be addressed by the proposed MTL adaptation with the auxiliary letter branch where the letter targets can be fully observed in adaptation data. 

The results of supervised MTL adaptation on the 7-speaker dataset are in the left sub-figure of Fig.~\ref{fig:MTL}, with different interpolation weights $\beta$ for letter targets. We find that $\beta=0.8$ gives WERs 16.98\%, 16.26\%, and 15.38\% for adaptation sets with 10, 50, and 200 utterances per speaker, respectively. The corresponding relative WERRs are 1.5\%, 5.7\%, and 10.8\% over the SI model. Above MTL results outperform those from KLD adaptation in Fig.~\ref{fig:KL}. In general, we obtain WER improvements with $\beta>0$. 

We then investigate unsupervised MTL adaptation for the word CTC model in the right sub-figure of Fig.~\ref{fig:MTL}. Following our description in Sec.~\ref{ssec:MTL}, we use the decoded hypothesis from the letter branch as the letter targets. Similar to our findings in supervised settings, $\beta=0.8$ also shows the best WER for unsupervised scenario. We obtain respectively 16.96\%, 16.58\%, and 16.17\% WERs when adapting with 10, 50, and 200 utterances, and correspondingly 1.6\%, 3.8\%, and 6.2\% relative WERRs over the SI model. Overall, MTL adaptation outperforms KLD regularization for both supervised and unsupervised scenarios. It is also interesting that the $\beta=1$, which completely relies on the letter targets to improve the word model, obtains similar WER as $\beta=0.8$. This indicates that the auxiliary letter task helps to adapt the CTC model towards the target speakers and addresses the output target sparsity issue with better generalization on unseen data.

\begin{table}[t]
  \caption{Comparison of WERs (\%) for various adaptation schemes using the word CTC mode on the 7-speaker set.}
  \label{tab:word_ctc_adapt_schemes}
  \centering
  \begin{tabular}{l|c|c|c|c|c}
    \toprule
    	\multirow{2}{*}{Adapt. type} & \multirow{2}{*}{$\#$ of utt.} & \multicolumn{2}{c|}{Supervised} & \multicolumn{2}{c}{Unsupervised} \\ \cline{3-6}
		 		&  & 	KLD		& 	MLT 	&	KLD	& 	MLT	\\
    \midrule
        Baseline &  - & \multicolumn{4}{c}{17.24} \\
    \midrule
  		& 10  & 18.04   & 17.83   & 17.60    & 17.99 \\ 
  Scalar      & 50  & 17.34   & 17.06   & 17.22   & 17.23 \\ 
        & 200 & 16.05   & 15.94   & 17.09   & 16.51 \\ \hline
  		& 10  & 17.17   & 17.08   & 17.27   & 17.08 \\ 
Linear        & 50  & 16.52   & 16.58   & 17.38   & 16.88 \\ 
        & 200 & 15.99   & 15.65   & 17.23   & 16.19 \\ \hline
  		& 10  & 17.18   & 16.98   & 17.24   & 16.96 \\ 
 Hidden       & 50  & 17.02   & 16.26   & 17.30   & 16.58 \\ 
        & 200 & 15.86   & 15.38   & 17.01   & 16.17 \\ 
    \bottomrule
  \end{tabular}
\end{table}

\subsubsection{Performance with Transformation-Based Adaptation}
Another widely-used approach to address the data sparsity issue is to insert SD linear transformation layers above certain layers of the SI model and optionally constrain the transformation in a structured form. Thus, the limited number of free parameters prevents adaptation from overfitting.
It is interesting to examine the impact of the transformation-based adaptation to the proposed conservative adaptation criteria. In this section, we evaluate two transformation-based adaptation schemes. First, we insert a linear layer above the second-to-last layer to transform the SI model. The linear adaptation reduces the SD footprint to around 1M parameters. Second, we apply element-wise linear transformation to the outputs of all the hidden layers (denoted as scalar), similar to learning hidden unit contribution (LHUC) \cite{swietojanski2016learning} and sigmoid adaptation \cite{zhao2015}. The scalar adaptation requires only 12K  adaptation parameters.
We set $\alpha=0$ and $\alpha=0.2$ for supervised and unsupervised KLD adaptation, respectively, and $\beta=0.8$ for both supervised and unsupervised MLT adaptation. These settings are also found to produce good results for the transformation-based adaptation. 

The performance of these systems in various scenarios on the 7-speaker set is shown in Table~\ref{tab:word_ctc_adapt_schemes}. 
In general, both scalar and linear adaptations exhibit similar WER patterns to directly adapting the hidden layers in all of the operating conditions. For supervised adaptation, all three adaptation schemes produce significant improvements over the SI model by 7-8\% on 200 development utterances (per speaker), even without regularization. The MTL adaptation further improves results by 2-3\% relative WER. For unsupervised adaptation, the KLD regularization barely improves the results, even for low footprint adaptation models. This indicates that simply restricting the amount of model parameters could not alleviate the target sparsity issue. When the MTL criterion is employed, all three adaptation schemes yield 4-6\% relative gain over the SI model on 200 development utterances. Both the linear adaptation and direct hidden layer adaptation benefit more from the MTL criterion than the scalar adaptation. 
In the following experiments, we will report results by directly updating the model parameters instead of inserting linear or scalar layers. 

Finally we extend our work to a larger 50-speaker dataset. We apply MTL adaptation with $\beta=0.8$, the best setting for the 7-speaker dataset for both supervised and unsupervised adaptations. We list the corresponding results in Table~\ref{tab:wer50}. The  MTL adaptation obtains 8.8\% and 4.0\% relative WERRs for the supervised and unsupervised scenarios, respectively.

\begin{table}[t]
  \caption{WERs and relative WERRs of supervised and unsupervised MTL adaptations for the word CTC model on the 50-speaker set with  $\beta=0.8$.}
  \label{tab:wer50}
  \centering
  \begin{tabular}{l|c|c|c}
    \toprule
								 		& $\#$ of utt. & 	WER (\%)		& 	relative WERR	(\%) \\
    \midrule
		Baseline						&		-				& 	13.88 & 	-	  \\	\hline
										&		10				& 	13.59 & 	2.1	 \\
		Supervised						&		50				& 	13.14 & 	5.3	\\
										&		200				& 	\textbf{12.66} & \textbf{8.8}	 \\	\hline
										&		10				& 	13.68 & 	1.4		\\
		Unsupervised					&		50				& 	13.44 & 	3.2		\\
										&		200				& 	\textbf{13.33} & \textbf{4.0}		\\	
    \bottomrule
  \end{tabular}
  \vspace{-4mm}
\end{table}
\subsection{Speaker Adaptation for CTC with Mix-unit Outputs}
We also apply both the KLD and MTL adaptations for the mix-unit CTC model. The baseline WERs of SI mix-unit CTC model are 16.76\% on the 7-speaker dataset and 13.63\% on the 50-speaker dataset, respectively.
\vspace{-3mm}
\subsubsection{KLD Adaptation}
\vspace{-1mm}
We first evaluate the KLD adaptation for the mix-unit CTC model on the 7-speaker dataset. The results from the supervised KLD adaptation are in Table~\ref{tab:mix-unit-kld-sup}. Similar to the word CTC model, we observe consistent WER improvements with increasing number of adaptation utterances per speaker as expected. The best WER (15.48\%) is obtained from adapting all hidden layers with $\alpha=0.0$ on 200 utterances per speaker, providing a relative WERR 7.63\% over the baseline SI mix-unit CTC model. We also observe the diminishing benefit of larger regularization for larger dataset (200 utterances) for adapting hidden layers. 

Table~\ref{tab:mix-unit-kld-uns} shows the results of unsupervised KLD adaptation for the mix-unit model. Consistent with the supervised scenario, adapting hidden layers provides the best WER (16.14\%) and the relative WERR is 3.70\% on 200 utterance with $\alpha=0.0$. In general,  adapting all hidden layers performs the best.

\begin{table}[t]
  \caption{WERs (\%) for supervised KLD regularization adaptation for the mix-unit CTC model on the 7-speaker set.}
  \label{tab:mix-unit-kld-sup}
  \centering
  \begin{tabular}{l|c|c|c|c|c}
    \toprule
		Adapt. type 		& $\#$ of utt. & 	$\alpha$=0		& 	$\alpha$=0.2 	&		$\alpha$=0.5 	& 	$\alpha$=0.8	\\
   	\midrule
        Baseline &  - & \multicolumn{4}{c}{16.76} \\
    \midrule
										&		10				& 	17.48 & 17.07	& 	16.98	& 16.91 \\
		All							&		50				& 	17.34 & 16.90 &  16.76	& 16.76 \\
										&		200				& 16.63 & \textbf{15.73} & 	16.02	& 16.49 \\	\hline
										&		10				& 	16.89 & 	16.80		& 	16.80	& 16.76 \\
		Hidden					&		50				& 	16.43 &  16.43	& 	16.59	& 16.69 \\
										&		200				& 	\textbf{15.48} & 15.83	& 	16.15	& 16.78 \\	
    \bottomrule
  \end{tabular}
\end{table}
\vspace{-2mm}

\begin{table}[t]
  \caption{WERs (\%) for unsupervised KLD regularization adaptation for the mix-unit CTC model on the 7-speaker set.}
  \label{tab:mix-unit-kld-uns}
  \centering
  \begin{tabular}{l|c|c|c|c|c}
    \toprule
		Adapt. type 		& $\#$ of utt. & 	$\alpha$=0		& 	$\alpha$=0.2 	&		$\alpha$=0.5 	& 	$\alpha$=0.8	\\
   	\midrule
        Baseline &  - & \multicolumn{4}{c}{16.76} \\
    \midrule
										&		10				& 	17.09 & 	16.87	& 	16.80	& 16.83 \\
		All							&		50				& 	16.99 &  16.70	& 	16.78	& 16.84 \\
										&		200				& 16.51 & \textbf{16.40} & 	16.53	& 16.83 \\	\hline
										&		10				& 	16.71 &  16.73	& 	16.76	& 16.83 \\
		Hidden					&		50				& 	16.45 & 16.60	& 	16.68	& 16.75 \\
										&		200				& 	\textbf{16.14} & 16.24	& 	16.61	& 16.67 \\	
    \bottomrule
  \end{tabular}
  \vspace{-2mm}
\end{table}
\vspace{-2mm}

\subsubsection{MTL Adaptation}
\vspace{-1mm}
In Table~\ref{tab:mix-unit-mtl}, we report the results of supervised and unsupervised MTL adaptations for the mix-unit CTC model on the 7-speaker dataset. For the supervised MTL adaptation, the best WERs are 16.60\%, 15.88\%, and 14.96\% for adapting with 10, 50, 200 utterances, respectively. The corresponding relative WERRs are 0.97\%, 5.25\%, and 10.75\% over the baseline SI model. For the unsupervised MTL adaptation, we observe best WERs as 16.53\%, 15.92\%, and 15.63\% with 10, 50, 200 adaptation utterances, respectively. The corresponding relative WERRs are 1.37\%, 5.04\%, and  6.79\% over the baseline SI model. Both the best relative WERRs for supervised and unsupervised scenarios by MTL adaptations are larger than the best WERRs by KLD adaptations. This indicate MTL adaptation is more effective than KLD adaptation for the mix-unit CTC model.

\begin{table}[t]
  \caption{WERs (\%) for supervised and unsupervised MTL adaptations for the mix-unit CTC model on the 7-speaker set.}
  \label{tab:mix-unit-mtl}
  \centering
  \begin{tabular}{l|c|c|c|c|c}
    \toprule
		Adapt. type 		& $\#$ of utt. & 	$\beta$=1		& 	$\beta$=0.8 	&		$\beta$=0.5 	& 	$\beta$=0.2	\\
   	\midrule
        Baseline &  - & \multicolumn{4}{c}{16.76} \\
    \midrule
										&		10				& 	16.81 & 16.60	& 	16.67	& 16.63 \\
		Supervised							&		50				& 	16.17 & 16.10 &  15.97	& 15.88 \\
										&		200				& 15.08 & \textbf{14.96} & 15.13	& 15.28 \\	\hline
										&		10				& 	16.62 & 	16.53		& 	16.64	& 16.61 \\
		Unsupervised					&		50				& 	16.16 &  16.06	& 	15.94	& 15.92 \\
										&		200				& 	15.82 & 15.67	& 	\textbf{15.63}	& 15.66 \\											
    \bottomrule
  \end{tabular}
  \vspace{-2mm}
\end{table}

Finally, we present the results of supervised and unsupervised MTL adaptations on the 50-speaker dataset in Table~\ref{tab:mix-unit-50spk} by using the best setting for the 7-speaker dataset: $\beta=0.5$ and $\beta=0.8$ for supervised and unsupervised MTL adaptations.  We obtain the best relative WERRs as respectively 9.6\% and 3.8\% for supervised and unsupervised adaptations over the baseline SI mix-unit CTC model.
\vspace{-4mm}
\begin{table}[t]
  \caption{WERs and relative WERRs of supervised and unsupervised MTL adaptations for a mix-unit CTC model on the 50-speaker set with $\beta=0.5$ and $\beta=0.8$, respectively.}
  \label{tab:mix-unit-50spk}
  \centering
  \begin{tabular}{l|c|c|c}
    \toprule
								 		& $\#$ of utt. & 	WER (\%)		& 	relative WERR	(\%) \\
    \midrule
		Baseline						&		-				& 	13.63 & 	-	  \\	\hline
										&		10				& 	13.40 & 	1.7	 \\
		Supervised						&		50				& 	12.95 & 	5.0	\\
										&		200				& 	\textbf{12.32} & \textbf{9.6}	 \\	\hline
										&		10				& 	13.54 & 	0.7		\\
		Unsupervised					&		50				& 	13.31 & 	2.3		\\
										&		200				& 	\textbf{13.11} & \textbf{3.8}		\\	
    \bottomrule
  \end{tabular}
  \vspace{-2mm}
\end{table}

\section{Conclusions}
\label{sec:con}
In this study, we propose KLD regularization and MTL for speaker adaptation of three types of E2E CTC models using letters, words, and mix-units as output targets respectively. 
The KLD adaptation constrains the adapted model to be close to the SI model to mitigate overfitting the adaptation set. To better address the output target sparsity issue when the CTC models have tens of thousands of output targets, we developed MTL adaptation which has an auxiliary task of predicting letter targets that can be fully observed despite limited adaptation data. 

We evaluated the two proposed methods on the Microsoft SMD task in both supervised and unsupervised settings. Our experiments demonstrate that the KLD regularization is useful for the three CTC models. The large number of word and mix-unit CTC outputs, most of which are unseen in limited speaker adaptation data, adds additional challenge to adaptation. 
Including an auxiliary task of predicting letters in MTL adaptation better addresses the output target sparsity issue and further improves the adaptation performance.
The MTL adaptation improves the baseline SI word CTC model by up to 8.8\% and 4.0\% relative WERRs for supervised and unsupervised adaptation, and obtains up to 9.6\% and 3.8\% WERRs over the baseline SI mix-unit CTC model, respectively. Results show that the MTL adaptation performs better than KLD adaptation for both word and mix-unit CTC models. This study presents our first investigation of speaker adaptation for E2E CTC models. In future work, we expect to apply the proposed methods to adapt other E2E models, including the attention based encoder-decoder models and the RNN transducers.

\bibliographystyle{IEEEbib}
\bibliography{strings,refs}

\end{document}